%% file: acl_latex.tex
\documentclass[11pt]{article}

\usepackage[final]{acl}

\usepackage{times}
\usepackage{latexsym}

\usepackage[T1]{fontenc}

\usepackage[utf8]{inputenc}

\usepackage{microtype}

\usepackage{inconsolata}

\usepackage{graphicx}

\usepackage{times}
\usepackage{latexsym}
\usepackage[T1]{fontenc}
\usepackage[utf8]{inputenc}
\usepackage{microtype}
\usepackage{inconsolata}

\usepackage{algorithm}
\usepackage{algorithmic}
\usepackage{amsmath}
\usepackage{xcolor}

\usepackage{amsmath, amsfonts}  
\usepackage{multirow}
\usepackage{svg}
\usepackage{caption}
\usepackage{subcaption}
\usepackage{wrapfig}
\usepackage{enumitem}
\usepackage{amssymb}
\usepackage{mathtools}
\usepackage{framed}

\definecolor{darkgreen}{rgb}{0.0, 0.3, 0.0}

\usepackage{newfloat}
\usepackage{listings}

\usepackage{tcolorbox}
\usepackage{booktabs}
\tcbuselibrary{skins, breakable}

\usepackage{microtype}
\usepackage{hyperref}
\usepackage{url}
\usepackage{booktabs}

\usepackage{lineno}

\definecolor{darkblue}{rgb}{0, 0, 0.5}
\hypersetup{colorlinks=true, citecolor=darkblue, linkcolor=darkblue, urlcolor=darkblue}

\title{DrugReason: Dynamic Multi-View Reasoning over Knowledge Graph and Language Evidence for Drug Repurposing}

\author{
  \textbf{Zijie Liu}$^{1,*}$ \quad
  \textbf{Hongxuan Li}$^{1,*}$ \quad
  \textbf{Zhen Tan}$^{2}$ \quad
  \textbf{Jinhao Duan}$^{1}$ \\
  \textbf{Baixiang Huang}$^{3}$ \quad
  \textbf{Zunpeng Liu}$^{4}$ \quad
  \textbf{Kai Shu}$^{3}$ \quad
  \textbf{Tianlong Chen}$^{1,\dagger}$ \\
  $^{1}$University of North Carolina at Chapel Hill \\
  $^{2}$Stevens Institute of Technology \\
  $^{3}$Emory University \\
  $^{4}$Massachusetts Institute of Technology \\
  $^{*}$Equal contribution. \\
  $^{\dagger}$Corresponding author: \texttt{tianlong@cs.unc.edu}
}

\begin{document}
\maketitle
\begin{abstract}
Drug repurposing aims to identify new therapeutic uses for existing compounds and, compared with de novo drug discovery, offers a faster and more cost-effective path to clinical translation. However, the space of candidate drug–disease pairs is enormous and their underlying relationships often depend on complex multi-hop biological mechanisms, making it difficult to reliably predict which pairs represent true therapeutic relationships. Existing approaches tackle this from two directions: knowledge graph-based methods organize curated biomedical evidence into structured relational networks for grounded multi-hop reasoning, while LLM-based methods leverage pretrained knowledge to generate flexible mechanistic rationales. Yet neither is sufficient alone — KGs are confined to observed graph structure while LLMs lack factual grounding and risk hallucination. To address this gap, we propose DrugReason, a multi-view reasoning framework that integrates grounded KG reasoning with LLM-generated mechanistic inference for drug repurposing. DrugReason adaptively routes diverse reasoning paths to specialized experts conditioned on the query context, while a cross-expert distillation objective enables knowledge sharing without sacrificing expert specialization. Experiments on PharmaDB, DDInter, and DrugBank show that
\textsc{DrugReason} improves average performance over strong single-view reasoning baselines and achieves competitive or superior results compared with graph-based alternatives, while providing interpretable routing-based predictions.
\end{abstract}

\section{Introduction}
\label{sec:intro}
\input{sections/intro}

\section{Related Work}
\label{sec:related_work}
\input{sections/related_work}

\section{Method}
\label{sec:method}
\input{sections/method}

\section{Experiments}
\label{sec:exp}
\input{sections/exp}

\section{Conclusion}
\label{sec:conclusion}
We presented \textsc{DrugReason}, a dynamic multi-view reasoning framework
for drug-repurposing-oriented biomedical relation prediction. It integrates
KG-derived factual grounding with training-time LLM-generated mechanistic
rationales, adaptively routes heterogeneous reasoning views to two
specialized generative experts, and applies cross-expert mutual
distillation to share predictive signals while preserving specialization.
Experiments on DDInter, DrugBank, and PharmaDB show that explicit reasoning
evidence is critical, single-view and static-fusion strategies are often
unstable, and \textsc{DrugReason} achieves strong overall performance
against reasoning, graph-based, and transformer baselines. These results
highlight heterogeneous, query-dependent evidence interpretation as a
promising direction for reliable and interpretable drug repurposing.

\section*{Limitations}
\textsc{DrugReason} is intended as a computational reasoning framework for prioritizing candidate relations, not as experimental validation of therapeutic efficacy or safety. Its performance depends on the coverage and quality of the underlying biomedical KG, as missing or noisy graph relations can affect the constructed reasoning views. The LLM-generated views may also contain unsupported mechanistic claims, although routing and mutual distillation help reduce reliance on any single view. In addition, using two generative LLM experts increases computational cost compared with single-model prompting or graph-only baselines. Future work will explore more efficient routing, stronger automatic rationale verification, and broader evaluation on prospective drug repurposing settings.

\section*{Acknowledgments}
This research was partially supported by the National Science Foundation (NSF) under Award No. IIS-2551752


\bibliography{custom}

\appendix

\label{sec:appendix}
\input{sections/app}

\end{document}

%% file: sections/intro.tex
\begin{figure}[t]
  \centering
  \includegraphics[width=1\columnwidth]{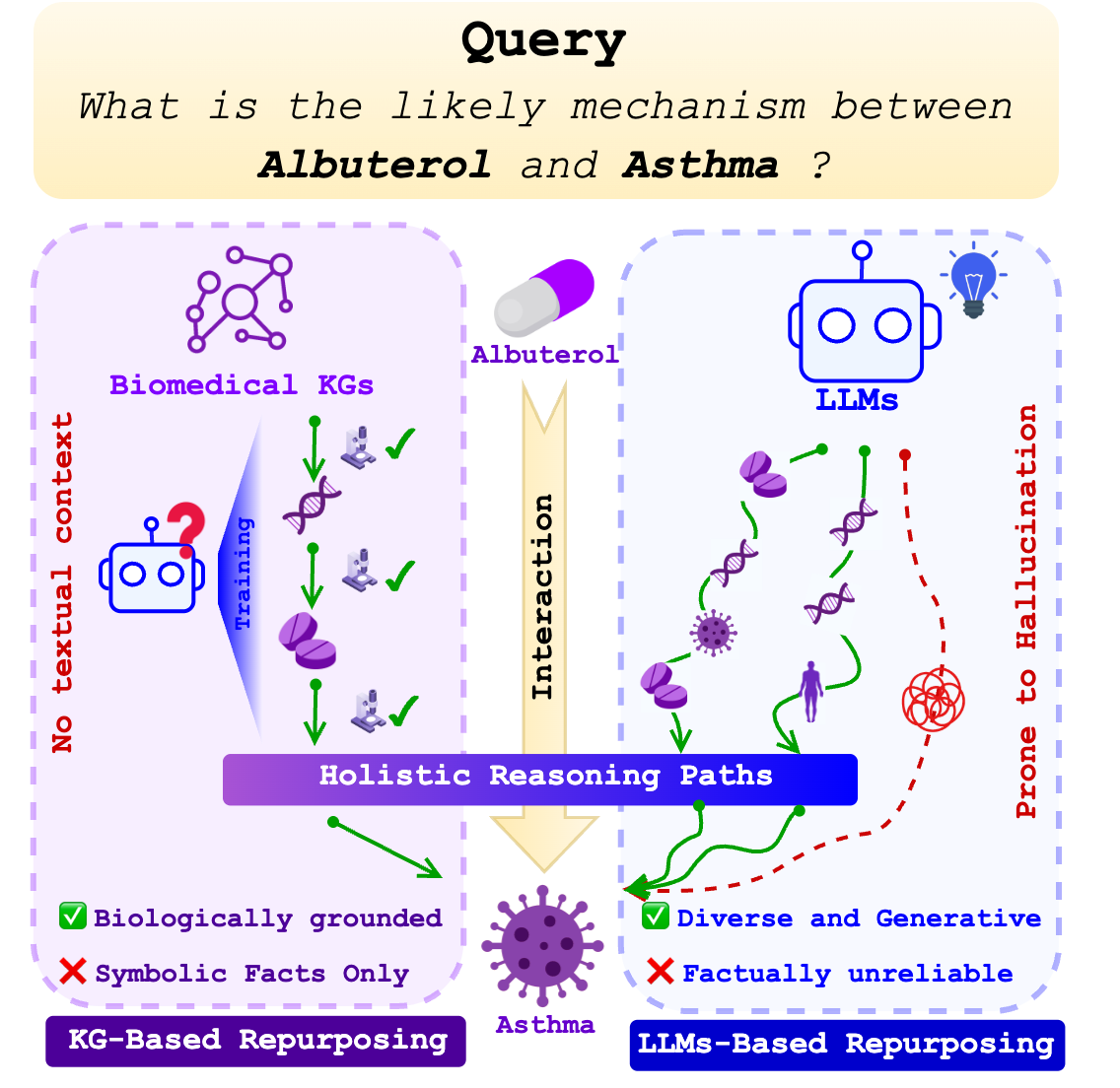}
  \caption{\small \textbf{Motivation for evidence-adaptive multi-view reasoning.}
    KG-based methods provide grounded biomedical evidence through curated relational structure, while LLM-based methods generate flexible mechanistic hypotheses but may hallucinate without factual grounding.
    DrugReason combines KG-derived and LLM-generated reasoning views and adaptively routes them according to the query context for drug repurposing.}
  \label{fig:motivation}
\end{figure}

Drug repurposing aims to identify new therapeutic uses for existing compounds, offering a faster and more cost-effective alternative to de novo drug discovery~\cite{pushpakom2019drug,hopkins2008network}. Despite its promise, computational drug repurposing remains challenging because candidate drug--disease relationships often depend on complex biological mechanisms involving drugs, genes, proteins, pathways, and diseases~\cite{hopkins2008network,himmelstein2017systematic,chandak2023building}. These mechanisms are rarely captured by a single direct association; instead, they often require reasoning over diverse biomedical evidence and multi-hop biological connections. In this paper, we study biomedical relation prediction over drug-involving entity pairs, encompassing both drug--disease therapeutic associations and drug--drug interactions.

Existing approaches address this problem from two directions. Knowledge graph (KG)-based methods organize curated biomedical entities and relations into structured networks, enabling grounded reasoning over multi-hop paths between drugs, diseases, genes, and pathways~\cite{himmelstein2017systematic,chandak2023building,perdomo2024knowledge}. Such graph structure provides interpretable evidence for repurposing relations, especially when relevant biological routes are explicitly represented. In parallel, large language model (LLM)-based methods leverage pretrained knowledge to generate natural-language mechanistic rationales, such as causal chains, pathway-level explanations and alternative biological hypotheses~\cite{singhal2023large,chen2025benchmarking,liu2025generalist}. 

However, neither direction is sufficient alone. KG-based methods are constrained by observed graph structure and relation coverage, while LLM-based methods may generate unsupported or hallucinated mechanisms without factual grounding~\cite{Ji_2023}. Recent work has therefore explored grounding LLMs with biomedical knowledge, retrieval, or graph-derived context~\cite{soman2024biomedicalknowledgegraphoptimizedprompt,cabello2025megmedicalknowledgeaugmentedlarge,li2024biomedragretrievalaugmentedlarge,sohn2025rationaleguidedretrievalaugmentedgeneration,abdullahi2025k,wei2024drugrealign}. Yet combining KG and LLM evidence does not resolve the core challenge.
Different reasoning paths encode different types of information: KG paths expose structured graph evidence, while CoT- and ToT-style rationales elicit sequential or hypothesis-level reasoning~\cite{abdullahi2025k,wei2022chain,yao2023tree}.
Static fusion strategies such as concatenation, voting, or fixed aggregation may not fully use such heterogeneous path-specific signals, a limitation also observed in multi-view and multimodal fusion settings~\cite{sahu2021adaptive,han2022trusted}.
This motivates an adaptive routing mechanism over diverse reasoning views.

To this end, we propose \textsc{DrugReason}, a multi-view reasoning framework for drug repurposing. For each candidate pair, \textsc{DrugReason} combines KG-derived evidence with LLM-generated mechanistic inference to construct multiple reasoning views. Rather than concatenating or averaging these views, \textsc{DrugReason} adaptively routes each view to a specialized expert based on the query and evidence content. Cross-expert distillation further shares transferable biomedical signals across routed experts while preserving their specialization. 
Experiments on PharmaDB, DDInter, and DrugBank show that DrugReason improves average performance over strong single-view reasoning baselines and achieves competitive or superior results compared with graph-based and static multi-view alternatives. Ablation studies show that adaptive routing and cross-expert mutual distillation both contribute to the overall gains, while routing distributions provide evidence-level transparency into how different reasoning views support each prediction. Our main contributions are as follows:

\begin{itemize}
    \item We introduce a multi-view reasoning formulation for drug repurposing that integrates KG-derived evidence and LLM-generated mechanistic rationales across drug–disease and drug–drug relation types.
    \item We propose \textsc{DrugReason}, which adaptively routes heterogeneous reasoning views to two specialized generative experts with cross-expert mutual distillation for knowledge sharing without collapsing specialization.
    \item Experiments on PharmaDB, DDInter, and DrugBank show strong improvements over single-view and static-fusion baselines, with ablations confirming the contribution of both routing and distillation.
\end{itemize}

%% file: sections/related_work.tex
\paragraph{Knowledge-Grounded Reasoning for Drug Discovery.}
Knowledge graph (KG)-based methods support drug repurposing by organizing curated biomedical entities and relations into structured networks~\cite{sys_kg,Rephetio,kg_review}. 
Path-based methods provide interpretable drug--disease mechanisms~\cite{k-path,silico}, while GNN-based methods learn expressive entity representations but are often less transparent and remain bounded by observed graph structure~\cite{hgtdr,DRMAHGC,TxGNN}. 
Recent KG-to-LLM approaches, including K-Paths, DrugReAlign, and graph2prompt, verbalize KG paths as prompts to improve factual grounding~\cite{k-path,DrugReAlign,graph2prompt}. 
However, most of them use KG evidence as static context rather than adaptively deciding how different evidence views should be interpreted for each query.

\paragraph{LLM Biomedical Reasoning.}
Large language models and biomedical foundation models have shown strong potential on medical and biomedical reasoning tasks~\cite{Med-PaLM,llm_med_benchmark,BioGPT,PubMedGPT,MedFound}. 
In drug discovery, systems such as DrugAgent, MedReason, Tx-LLM, DrugMCTS, and PharmAgents combine domain knowledge, external tools, or agent collaboration to support interpretable therapeutic prediction~\cite{DrugAgent,MedReason,Tx-LLM,DrugMCTS,PharmAgents2025}. 
These methods broaden mechanistic reasoning, but generated rationales can still be unreliable without structured grounding and evidence-specific control.
More broadly, evaluations of generative models have examined relevance-aware uncertainty in free-form LLM outputs, strategic LLM reasoning, and membership leakage in diffusion models~\cite{duan2024shifting,duan2024gtbench,duan2023diffusion}. These complementary reliability dimensions motivate careful handling of generated evidence.

\paragraph{Diverse LLM Reasoning Strategies.}
Reasoning prompts such as Chain-of-Thought and Tree-of-Thought encourage LLMs to decompose complex questions and explore alternative hypotheses~\cite{wei2022chain,yao2023tree}. 
Such strategies increase reasoning diversity, including in biomedical hypothesis generation, but they typically treat generated paths uniformly. 
In contrast, \textsc{DrugReason} treats KG-derived and LLM-generated rationales as heterogeneous reasoning views and routes each view to a specialized expert.

%% file: sections/method.tex
\begin{figure*}[!t]
  \centering
  \includegraphics[width=\textwidth]{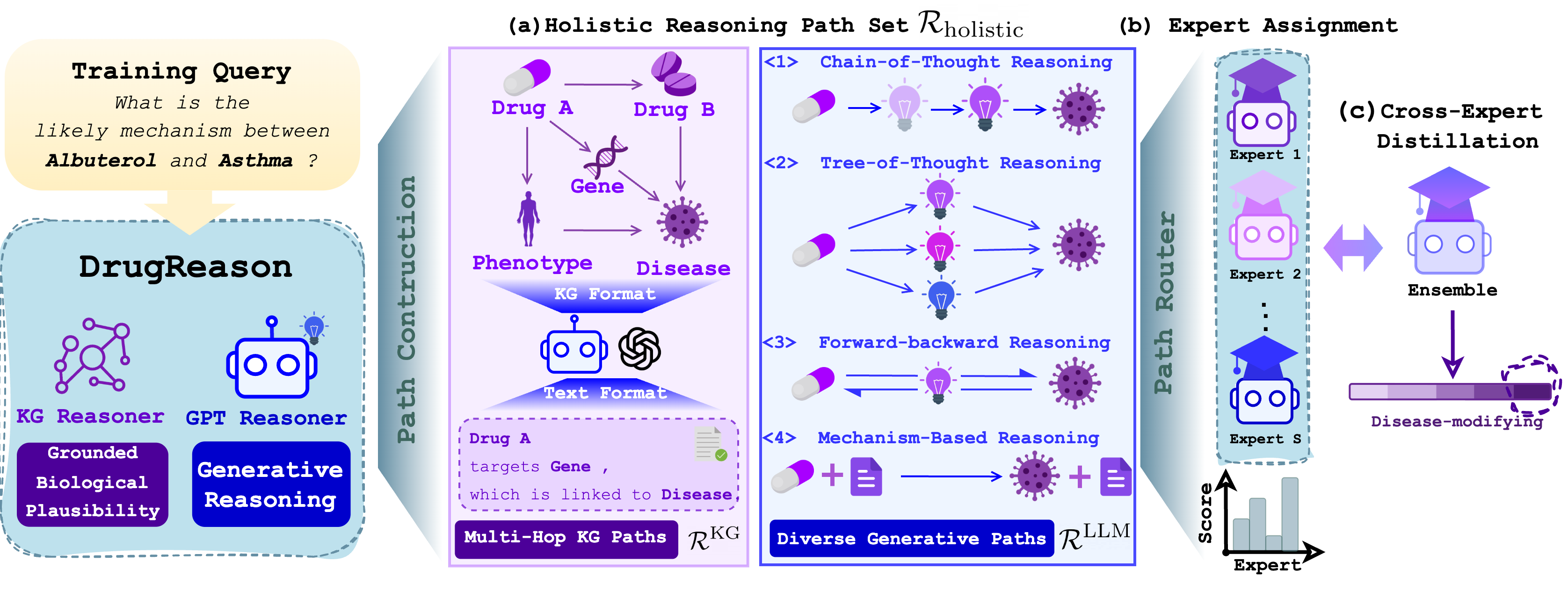}
  \caption{\small
  \textbf{Overview of \textsc{DrugReason}.}
  Given a candidate pair, \textsc{DrugReason} constructs multiple reasoning views from KG-derived evidence and LLM-generated mechanistic inference. Each view is serialized with the query context and encoded for adaptive routing. A conditional router assigns each view to one of two generative LLM experts, Llama or Qwen, which predicts a relation label from the routed view. During training, cross-expert mutual distillation exchanges softened label distributions between the two routed experts while preserving their specialization. At inference, routed view-level predictions are aggregated to produce the final relation label.}
  \label{fig:pipeline}
\end{figure*}

\subsection{Problem Formulation and Overview}

Given a candidate pair $(u,v)$, where $u$ is a drug and $v$ is either a disease or another drug depending on the benchmark, the goal is to predict a relation label $y \in \mathcal{Y}$. For drug--disease pairs, $y$ denotes a therapeutic association label; for drug--drug pairs, $y$ denotes a drug--drug relation label. Since our experts are generative LLMs, each label $y$ is represented by a canonical label verbalizer $\nu(y)$, i.e., the text sequence that the model should generate for that relation label.

For each pair, \textsc{DrugReason} constructs a set of reasoning views
\begin{equation}
\mathcal{R}(u,v)=\{r_1,\ldots,r_m\},
\end{equation}
where each $r_j$ is a serialized textual evidence instance generated from one reasoning strategy. For KG-derived views, $r_j$ is an entity definition or a verbalized graph path; for LLM-generated views, $r_j$ is a mechanistic rationale generated under a specific prompt. Thus, all evidence sources are represented as text while preserving their different reasoning roles.

\textsc{DrugReason} proceeds in three stages. First, it constructs $\mathcal{R}(u,v)$ by combining KG-derived evidence with LLM-generated mechanistic inference. Second, each reasoning view is serialized with the candidate query and routed to one of two generative LLM experts. The selected expert predicts a relation label for that view. Third, routed view-level predictions are aggregated into the final pair-level prediction. During training, we further apply symmetric mutual distillation between the two LLM experts so that they can exchange softened predictive signals without removing routed specialization.

\subsection{Reasoning View Construction}

For each candidate pair $(u,v)$, \textsc{DrugReason} constructs reasoning views that describe the same query from different evidence sources. We use the term \emph{reasoning view} to denote a serialized textual evidence instance generated from one reasoning strategy. KG-derived views provide structured grounding from curated biomedical resources, while LLM-generated views provide mechanistic inference in natural language. All views are serialized into a unified text format so that they can be routed and processed under the same framework.

\textbf{KG-derived views.}
The definition view provides semantic grounding for the entities in the query. We retrieve biomedical definitions for the drug and disease/drug entities from PrimeKG~\cite{chandak2023building}. These definitions help identify entity type, biological role, and pharmacological context before relation prediction.

The KG-path view provides graph-grounded multi-hop evidence between the two entities. We extract relational paths from the biomedical KG using K-Paths~\cite{abdullahi2025k}, which retrieves concise and diverse loopless paths between query entities. A path is represented as
\begin{equation}
P_j = (u, \rho_1, v_1, \rho_2, \ldots, \rho_l, v), \quad l \leq l_{\max},
\end{equation}
where $\rho_i$ denotes a KG relation and $v_i$ denotes an intermediate biomedical entity. Each extracted path is verbalized into text and used as one KG-derived reasoning view. We denote the KG-derived view set as $\mathcal{R}^{\mathrm{KG}}(u,v)$.

\textbf{LLM-generated views.}
KG-derived views are grounded in curated graph structure, but they may not fully express the mechanism behind a candidate relation. During training, we use a teacher LLM to construct additional mechanistic reasoning views under four prompting strategies. For LLM-generated views, we use four prompting strategies that capture different forms of mechanistic inference: chain-of-thought reasoning describes a sequential causal explanation connecting the candidate pair; tree-of-thought reasoning explores and compares multiple biological hypotheses; forward--backward validation reasons from $u$ to $v$ and from $v$ back to $u$ to check directional consistency; and mechanism-focused reasoning emphasizes molecular targets, pathways, pharmacological processes, and therapeutic mechanisms. These teacher-generated views are used only as training-time evidence for routing and expert learning. \textsc{DrugReason} does not query the teacher LLM at inference. The prompt templates and filtering procedure are provided in Appendix~\ref{app:prompt_filtering}.

The LLM-generated views form $\mathcal{R}^{\mathrm{LLM}}(u,v)$. The full reasoning-view set is
\begin{equation}
\mathcal{R}(u,v)=\mathcal{R}^{\mathrm{KG}}(u,v)\cup \mathcal{R}^{\mathrm{LLM}}(u,v).
\end{equation}

\textbf{View serialization.}
Each reasoning view is serialized with the candidate query into a unified input:
\begin{equation}
x_j = [q(u,v); r_j],
\end{equation}
where $q(u,v)$ is a short textual description of the candidate pair and $r_j$ is the corresponding reasoning view. We apply only deterministic formatting checks to remove empty outputs, malformed responses, and template artifacts such as explicit label fields. This step does not use human annotation and does not rewrite retained views.

\subsection{Adaptive Routing to Generative LLM Experts}

The reasoning views in $\mathcal{R}(u,v)$ encode different types of information and have different textual formats. A KG path is concise and relation-centered, while a mechanism-focused rationale is longer and narrative. Instead of concatenating all views or sending every view to the same expert, \textsc{DrugReason} first routes each reasoning view to a specialized generative LLM expert and then aggregates the routed label distributions. In our implementation, the expert set is
\begin{equation}
\mathcal{E}=\{E^{(1)},E^{(2)}\},
\end{equation}
where $E^{(1)}$ and $E^{(2)}$ correspond to the Llama and Qwen experts, respectively. The two experts remain separate throughout training and inference; routing determines which expert scores each view, while mutual distillation later exchanges only soft predictive distributions. Routing induces specialization by allowing each expert to learn from the views assigned to it.

For each serialized reasoning view $x_j=[q(u,v);r_j]$, we compute a routing representation
\begin{equation}
h_j = \mathrm{Enc}(x_j) \in \mathbb{R}^{d},
\end{equation}
where $\mathrm{Enc}(\cdot)$ is a lightweight text encoder used only for routing. The router maps $h_j$ to expert logits:
\begin{equation}
z_j = \mathrm{MLP}(h_j) \in \mathbb{R}^{S},
\end{equation}
where $S=2$ in our implementation. We use a straight-through Gumbel--Softmax router:
\begin{equation}
\tilde{\boldsymbol{\alpha}}_j =
\mathrm{softmax}\left((z_j+\mathbf{g})/\tau\right),
\,
g^{(s)} \sim \mathrm{Gumbel}(0,1),
\end{equation}
and take the hard assignment
\begin{equation}
\boldsymbol{\alpha}_j
=
\mathbf{e}_{\arg\max_{s\in[S]}\tilde{\alpha}^{(s)}_j}.
\end{equation}
Here, $\boldsymbol{\alpha}_j$ is one-hot in the forward pass, while gradients are propagated through the soft routing probabilities $\tilde{\boldsymbol{\alpha}}_j$.

Because the experts are generative LLMs, each expert predicts a relation label by scoring canonical label verbalizers. Let $\nu(y)=(w_1,\ldots,w_{|\nu(y)|})$ denote the token sequence for label $y$. For expert $E^{(s)}$, we compute the generative score of label $y$ under input $x_j$ as
\begin{equation}
a_{j,y}^{(s)}
=
\frac{1}{|\nu(y)|}
\sum_{t=1}^{|\nu(y)|}
\log P_{E^{(s)}}\!\left(w_t \mid x_j, w_{<t}\right),
\end{equation}
where length normalization avoids favoring shorter label strings. The view-level label distribution of expert $s$ is
\begin{equation}
p_j^{(s)}(y)
=
\frac{\exp(a_{j,y}^{(s)})}
{\sum_{y'\in\mathcal{Y}}\exp(a_{j,y'}^{(s)})}.
\end{equation}

The routed view-level prediction is
\begin{equation}
p_j
=
\sum_{s=1}^{S}
\alpha_j^{(s)} p_j^{(s)}.
\end{equation}
The pair-level prediction aggregates the routed predictions across all reasoning views:
\begin{equation}
\hat{p}
=
\frac{1}{|\mathcal{R}(u,v)|}
\sum_{j=1}^{|\mathcal{R}(u,v)|} p_j,
\quad
\hat{y}=\arg\max_{y\in\mathcal{Y}}\hat{p}_y .
\end{equation}

\subsection{Cross-Expert Mutual Distillation}

Adaptive routing encourages the two LLM experts to specialize on different reasoning views. In the first training stage, the router assigns reasoning views to experts, allowing each model to learn from the subset of views routed to it. However, this routed specialization can also separate useful information across experts. For example, one expert may learn stronger patterns from graph-grounded views, while the other may better capture signals from generated mechanistic rationales.

To reduce this separation without removing specialization, we apply symmetric mutual distillation in the second training stage. During this stage, both experts are evaluated on each serialized view to compute the mutual distillation loss, while the supervised prediction loss remains routed according to $\boldsymbol{\alpha}_j$. For each expert, we first compute temperature-scaled label distributions from the generative label scores:
\begin{equation}
p_{j,T}^{(s)}(y)
=
\frac{\exp(a_{j,y}^{(s)}/T)}
{\sum_{y'\in\mathcal{Y}}\exp(a_{j,y'}^{(s)}/T)},
\end{equation}
where $T$ is the distillation temperature.

We use a symmetric KL objective so that the two experts teach each other:
\begin{equation}
\begin{aligned}
\mathcal{L}_{\mathrm{mutual}}
&=
\frac{T^2}{2|\mathcal{R}|}
\sum_{j=1}^{|\mathcal{R}(u,v)|}
\Bigl[
\mathrm{KL}\!\left(
\mathrm{stopgrad}(p_{j,T}^{(1)}) \,\|\, p_{j,T}^{(2)}
\right)
\\
&\quad+
\mathrm{KL}\!\left(
\mathrm{stopgrad}(p_{j,T}^{(2)}) \,\|\, p_{j,T}^{(1)}
\right)
\Bigr].
\end{aligned}
\end{equation}
The stop-gradient operator prevents the target distribution in each KL term from being updated by the same term. Since the objective is bidirectional, each expert serves as both a soft teacher and a student through the two KL terms. This mutual distillation exchanges softened predictive information between the Llama and Qwen experts without sharing parameters. Because routed supervised learning is still maintained, the experts preserve their specialization while benefiting from each other's predictive distributions.

\subsection{Training Objective and Inference}

We train \textsc{DrugReason} in two stages. In Stage~1, the router assigns reasoning views to the two LLM experts, and each expert learns from the views routed to it. For a serialized reasoning view $x_j$, let $\alpha_j^{(s)} \in \{0,1\}$ denote the hard routing assignment to expert $s$. The routed supervised generation loss is
\begin{equation}
\begin{aligned}
\mathcal{L}_{\mathrm{pred}}
&=
-\frac{1}{|\mathcal{R}|}
\sum_{j=1}^{|\mathcal{R}|}
\sum_{s=1}^{2}
\alpha_j^{(s)}
\frac{1}{|\nu(y)|} \\
&\quad
\sum_{t=1}^{|\nu(y)|}
\log P_{E^{(s)}}\!\left(
w_t \mid x_j,w_{<t}
\right).
\end{aligned}
\end{equation}

Equivalently, this loss trains only the selected expert to generate the canonical label verbalizer for each routed reasoning view.

To avoid expert collapse while allowing confident view-level
assignments, we regularize the average routing distribution:
\begin{equation}
\bar{\alpha}^{(s)}
=
\frac{1}{|\mathcal{R}|}
\sum_{j=1}^{|\mathcal{R}|}
\tilde{\alpha}_j^{(s)},
\quad
\mathcal{H}_{\mathrm{load}}
=
-\sum_{s=1}^{2}
\bar{\alpha}^{(s)}
\log \bar{\alpha}^{(s)} .
\end{equation}
The Stage~1 objective is
\begin{equation}
\mathcal{L}^{(1)}
=
\mathcal{L}_{\mathrm{pred}}
-
\lambda_1 \mathcal{H}_{\mathrm{load}} .
\end{equation}

In Stage~2, we keep the routed supervised generation objective and add symmetric mutual distillation between the two LLM experts:
\begin{equation}
\mathcal{L}^{(2)}
=
\mathcal{L}_{\mathrm{pred}}
-
\lambda_1 \mathcal{H}_{\mathrm{load}}
+
\lambda_2 \mathcal{L}_{\mathrm{mutual}} .
\end{equation}
where $\lambda_1$ controls routing entropy regularization and $\lambda_2$ controls mutual distillation. The routed supervised loss preserves expert specialization, while the mutual distillation term allows the two experts to exchange soft predictive information.

At inference time, each reasoning view is encoded for routing, assigned to one expert, and then evaluated by the selected generative LLM expert. We use the selected expert's canonical label-verbalizer scores to obtain $p_j$, aggregate routed view-level distributions across all reasoning views, and output the final label:
\begin{equation}
\hat{p}
=
\frac{1}{|\mathcal{R}(u,v)|}
\sum_{j=1}^{|\mathcal{R}(u,v)|}
\sum_{s=1}^{2}
\alpha_j^{(s)} p_j^{(s)},
\,
\hat{y}=\arg\max_{y\in\mathcal{Y}}\hat{p}_y .
\end{equation}
The mutual distillation loss and routing entropy regularizer are used only during training.

%% file: sections/exp.tex
\paragraph{Experimental Setup.}
We evaluate \textsc{DrugReason} on three drug-repurposing-oriented
biomedical relation prediction benchmarks: DDInter~\cite{xiong2022ddinter}
and DrugBank~\cite{wishart2018drugbank} for drug--drug interaction
prediction, and PharmaDB~\cite{himmelstein2017systematic} for
drug--disease therapeutic association prediction. All experiments follow
an inductive protocol over Hetionet~\cite{sys_kg}, where test triples
contain entities unseen during task-specific training. We report accuracy
as the primary metric and Macro-F1 for additional control analyses.

We use \texttt{GPT-4o}~\cite{gpt4o} only as a training-time teacher to
generate mechanistic rationales. The two generative experts are
\texttt{Llama3.2-3B-Instruct}~\cite{llama3} and
\texttt{Qwen2.5-3B-Instruct}~\cite{qwen25}, with a lightweight
\texttt{RoBERTa-base} encoder~\cite{roberta} used only for routing
serialized reasoning views. We report expert-specific results after
routed training and mutual distillation, while both experts remain part
of the same \textsc{DrugReason} framework.

We compare with no-reasoning prompting, single-view reasoning baselines
including K-Path~\cite{k-path}, Chain-of-Thought~\cite{wei2022chain},
and Tree-of-Thought~\cite{yao2023tree}, static fusion controls, KGE
baselines including TransE~\cite{bordes2013translating} and
DistMult~\cite{yang2014embedding}, and graph/transformer baselines
including GCN~\cite{gcn}, GAT~\cite{gat}, GIN~\cite{gin},
GraphSAGE~\cite{hamilton2017inductive}, Graph Transformer~\cite{dwivedi2020generalization},
and Transformer~\cite{vaswani2017attention}. Additional implementation details and hyperparameters are provided in
Appendix~\ref{app:implementation}.

\begin{table*}[!t]
\centering
\renewcommand{\arraystretch}{1.1}
\resizebox{\textwidth}{!}{
\begin{tabular}{l|c|ccc|ccc}
\hline
\multirow{2}{*}{\textbf{Category}} & \multirow{2}{*}{\textbf{Prompt Type}} &
\multicolumn{3}{c|}{\texttt{Llama3.2}} &
\multicolumn{3}{c}{\texttt{Qwen2.5}} \\
& & \textbf{DDInter} & \textbf{DrugBank} & \textbf{PharmaDB} &
     \textbf{DDInter} & \textbf{DrugBank} & \textbf{PharmaDB} \\
\hline
No Reasoning & - & 0.20 & 17.30 & 43.25 & 3.34 & \underline{19.80} & 45.63 \\
\hline
\multirow{2}{*}{KG-only Reasoning}
& k-Path & 33.30 & \textbf{41.86} & 49.21 & 18.20 & 3.94 & 9.92 \\
& Definition & \underline{35.40} & \underline{36.01} & \underline{56.75} &
                \underline{63.40} & 16.55 & \underline{58.33} \\
\hline
\multirow{4}{*}{LLM-only Reasoning}
& Mechanism & 1.50 & 0.81 & 44.44 & 20.10 & 6.18 & 13.10 \\
& Chain of Thought & 20.40 & 14.42 & 45.24 & 11.80 & 1.55 & 0.79 \\
& Forward-Backward & -- & 15.91 & 53.17 & 20.30 & 0.85 & 1.59 \\
& Tree of Thought & 26.00 & 13.48 & 45.63 & 24.30 & 14.77 & -- \\
\hline
Our Method & - & \textbf{60.00} & 25.99 & \textbf{68.25} &
                \textbf{72.00} & \textbf{33.98} & \textbf{65.48} \\
\hline
\end{tabular}
}
\caption{\small
\textbf{Main comparison across reasoning strategies.}
KG-only and LLM-only views provide useful but uneven gains, while \textsc{DrugReason} achieves the best average performance across datasets and backbones by adaptively combining heterogeneous reasoning views.
}
\vspace{-5mm}
\label{tab:main_table}
\end{table*}

\textbf{RQ1: Why do we need adaptive multi-view reasoning?}
Table~\ref{tab:main_table} compares \textsc{DrugReason} with no-reasoning prompting and individual KG- or LLM-based reasoning views. Entity names alone are insufficient: on DDInter, no-reasoning prompting achieves only 0.20\% with \texttt{Llama3.2} and 3.34\% with \texttt{Qwen2.5}, while \textsc{DrugReason} reaches 60.00\% and 72.00\%. On PharmaDB, \textsc{DrugReason} also improves from 43.25\% to 68.25\% with \texttt{Llama3.2} and from 45.63\% to 65.48\% with \texttt{Qwen2.5}.

Single-view reasoning helps but is uneven. KG-derived views can be strong when graph evidence aligns with the target relation, e.g., Definition reaches 63.40\% on DDInter with \texttt{Qwen2.5}, and K-Path reaches 41.86\% on DrugBank with \texttt{Llama3.2}. However, no single view is reliable across all datasets and backbones, while LLM-only views are often unstable without structured grounding. By adaptively routing and aggregating heterogeneous views, \textsc{DrugReason} obtains the best result in five of six dataset--expert settings. The exception is DrugBank with \texttt{Llama3.2}, where K-Path remains stronger, suggesting that fine-grained DDI labels can sometimes favor direct graph-path evidence.

\begin{table}[t]
\centering
\small
\resizebox{\columnwidth}{!}{
\begin{tabular}{lccc}
\toprule
\textbf{Dataset} & \textbf{Best Single} & \textbf{Best Static} & \textbf{Best KGE} \\
\midrule
DDInter  & 34.00 / 23.36 & 20.40 / 8.61  & 43.00 / 30.58 \\
PharmaDB & 49.60 / 37.63 & 46.43 / 35.44 & 37.70 / 36.43 \\
\bottomrule
\end{tabular}
}
\caption{\small
\textbf{Control baselines for static fusion and graph embeddings.}
Each cell reports Acc./Macro-F1 (\%). Best Single denotes the best individual prompt; Best Static is selected by accuracy among concat-all, majority vote, and mean vote; Best KGE is selected by accuracy between TranSE and DistMult.
}
\vspace{-3mm}
\label{tab:control_compact}
\end{table}

\paragraph{RQ2: Is naive fusion enough?}
To rule out the possibility that \textsc{DrugReason} improves simply by using more evidence, we compare against prompt-based static fusion (concat-all, majority/mean voting) and KGE baselines (TransE, DistMult). As shown in Table~\ref{tab:control_compact}, static fusion is unreliable: on DDInter, the best single prompt achieves 34.00/23.36 while the best static fusion reaches only 20.40/8.61, and on PharmaDB static fusion still falls below the best single prompt. The KGE controls reveal a complementary limitation—TransE and DistMult are competitive on DDInter but weaker on PharmaDB, and on DrugBank the best KGE result is only 11.80/4.59, confirming that graph embeddings alone struggle with fine-grained 86-way prediction. Together, these controls support our core motivation: heterogeneous evidence must be interpreted adaptively rather than simply concatenated, voted, or embedded. Full static-fusion and KGE results are reported in Appendix~\ref{app:prompt_controls} and Appendix~\ref{app:kge_controls}.

\begin{figure*}[htbp]
\centering
\includegraphics[width=0.49\textwidth]{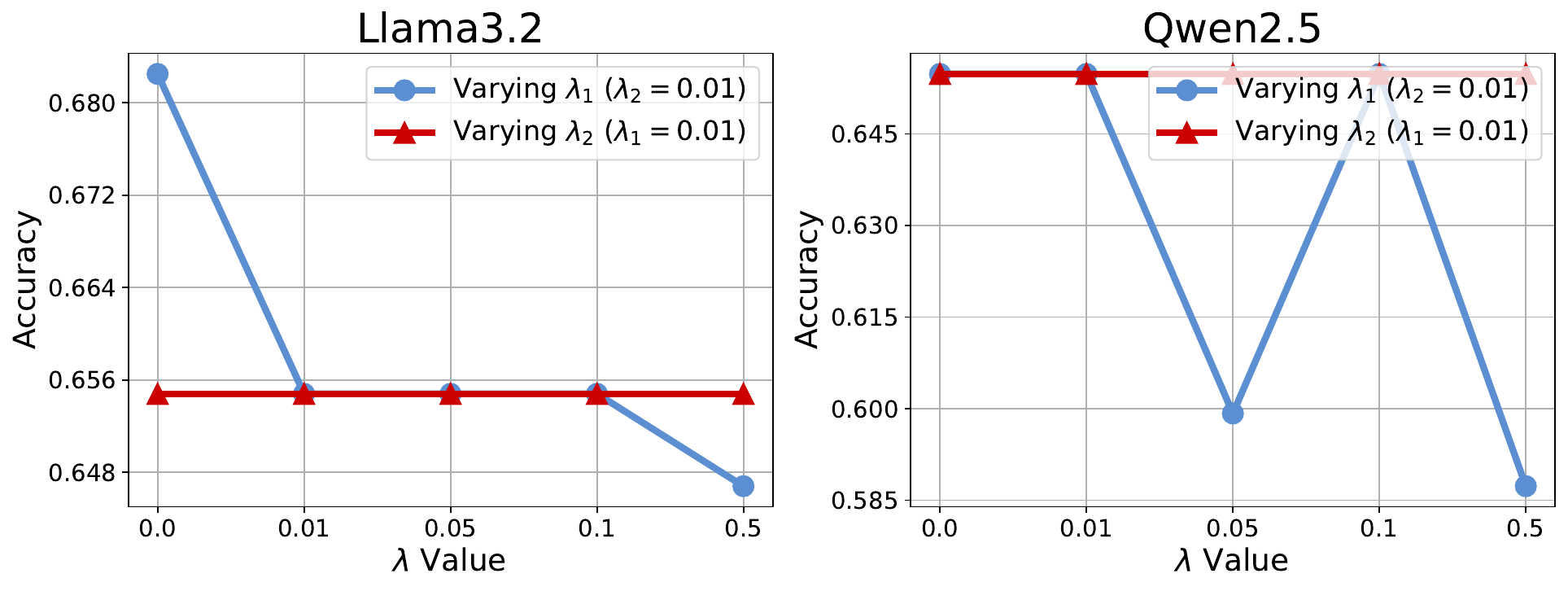}\hfill
\includegraphics[width=0.49\textwidth]{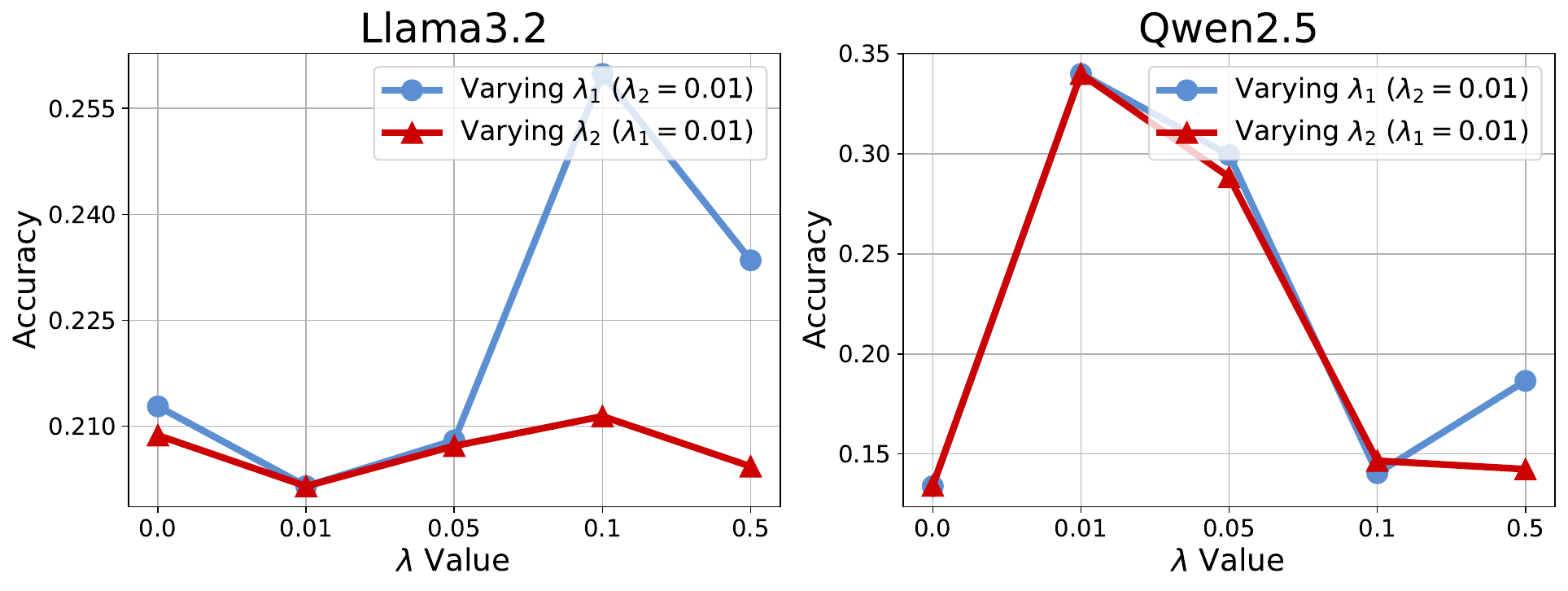}
\caption{\small
\textbf{Sensitivity of \textsc{DrugReason} to routing and distillation weights.} Varying $\lambda_1$ (routing regularization) has a larger impact than $\lambda_2$ (distillation weight) for both backbones, with performance generally degrading at larger values. \texttt{Llama3.2} remains relatively stable across most of the range, while \texttt{Qwen2.5} shows a sharper drop at $\lambda_1 = 0.05$ before partially recovering, indicating greater sensitivity to routing regularization strength.
}

\label{fig:hyper}
\end{figure*}

\begin{table}[t]
\centering
\small
\renewcommand{\arraystretch}{1.05}
\begin{tabular}{llcc}
\toprule
\textbf{Type} & \textbf{Model} & \textbf{DDInter} & \textbf{Phar.} \\
\midrule
\multirow{5}{*}{GNN}
 & GAT & 43.20 & \underline{67.11} \\
 & GCN & 41.60 & 65.35 \\
 & GIN & 25.40 & 47.81 \\
 & GraphSAGE & 60.70 & 62.72 \\
 & GraphTrans. & \underline{71.00} & 61.00 \\
\midrule
General & Transformer & 66.00 & 65.79 \\
\midrule
\multirow{2}{*}{Ours}
 & \texttt{Llama3.2} & 60.00 & \textbf{68.25} \\
 & \texttt{Qwen2.5} & \textbf{72.00} & 65.48 \\
\bottomrule
\end{tabular}
\caption{\small
\textbf{Comparison with graph-based and transformer baselines.}
Results are reported as accuracy (\%). The strongest \textsc{DrugReason} configuration is competitive with or better than graph and transformer baselines while producing language-level reasoning evidence.
}

\label{tab:gnn_compare}
\end{table}

\textbf{RQ3: How does \textsc{DrugReason} compare with graph-based baselines?}
We compare \textsc{DrugReason} with graph-based and transformer baselines to evaluate whether routed language reasoning can compete with models that directly operate on KG structure. As shown in Table~\ref{tab:gnn_compare}, \textsc{DrugReason} is competitive with or better than graph-based baselines on both tasks. On DDInter, \texttt{Qwen2.5} reaches 72.00\%, slightly exceeding GraphTrans. (71.00\%) and outperforming GAT, GCN, GIN, GraphSAGE, and the general Transformer. On PharmaDB, \texttt{Llama3.2} reaches 68.25\%, exceeding GAT (67.11\%). Performance is backbone-dependent—\texttt{Llama3.2} trails GraphTrans. on DDInter and \texttt{Qwen2.5} falls slightly below GAT on PharmaDB—so we do not claim universal dominance. The key result is that verbalized KG evidence and mechanistic language reasoning can match or exceed strong graph baselines while retaining interpretable language-level evidence for each prediction.

\begin{table}[t]
\centering
\small
\begin{tabular}{lccc}
\toprule
\multicolumn{4}{c}{\texttt{Llama3.2}} \\ \midrule
 & No Mut. & No Rtr. & Ours \\ \midrule
PharmaDB & \textbf{68.65} & 29.37 & 68.25 \\
DDInter  & 23.10 & 20.70 & \textbf{60.00} \\
DrugBank & 16.60 & 24.35 & \textbf{25.99} \\
\textbf{Avg.} & 36.12 & 24.81 & \textbf{51.41} \\ \midrule
\multicolumn{4}{c}{\texttt{Qwen2.5}} \\ \midrule
 & No Mut. & No Rtr. & Ours \\ \midrule
PharmaDB & 50.40 & 53.57 & \textbf{65.48} \\
DDInter  & 59.90 & 12.80 & \textbf{72.00} \\
DrugBank & 4.63 & 12.80 & \textbf{33.98} \\
\textbf{Avg.} & 38.31 & 26.39 & \textbf{57.15} \\ \bottomrule
\end{tabular}
\caption{\small
\textbf{Ablation study on mutual distillation (Mut.) and conditional routing (Rtr.).}
Results are reported as accuracy (\%). Removing routing causes large drops, and removing mutual distillation generally weakens average performance. The full model achieves the best average results, showing that routing and cross-expert knowledge sharing are complementary.
}

\label{tab:ablation_study}
\end{table}

\textbf{RQ4: Do routing and mutual distillation matter?}
Table~\ref{tab:ablation_study} isolates the effects of conditional routing
and cross-expert mutual distillation. Removing the router causes large
performance drops, especially on DDInter with \texttt{Qwen2.5}, where
accuracy falls from 72.00\% to 12.80\%. This confirms that heterogeneous
reasoning views should not be assigned uniformly. Mutual distillation
also improves average robustness by exchanging softened predictive
signals between the two experts. Its benefit is not monotonic in every
cell: on PharmaDB with \texttt{Llama3.2}, removing distillation slightly
improves accuracy from 68.25\% to 68.65\%. Overall, however, the full
model achieves the best average performance for both experts, showing
that routing and cross-expert knowledge sharing are complementary.

\begin{figure}[t]
  \centering
  \begin{subfigure}[b]{0.46\textwidth}
    \centering
    \includegraphics[width=\linewidth]{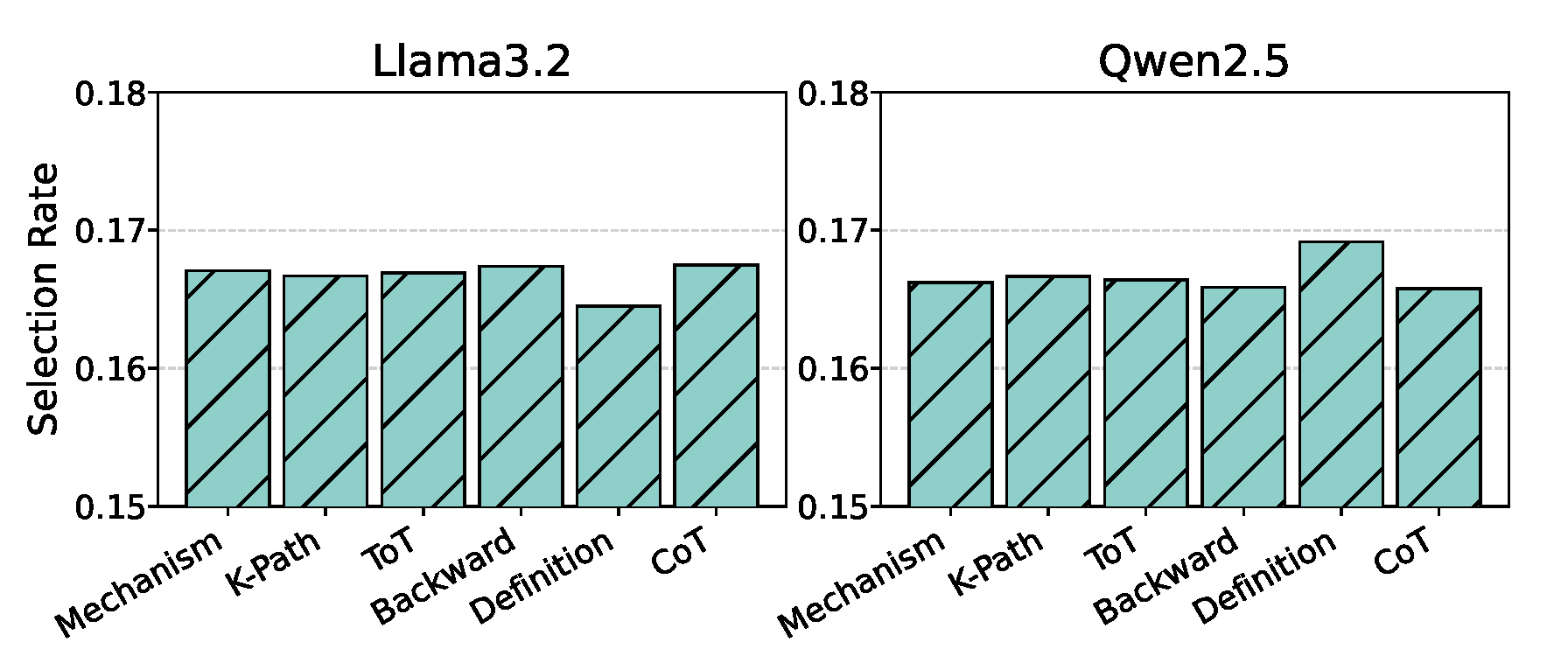}
    \caption{\small DDInter.}
  \end{subfigure}\\[1mm]
  \begin{subfigure}[b]{0.46\textwidth}
    \centering
    \includegraphics[width=\linewidth]{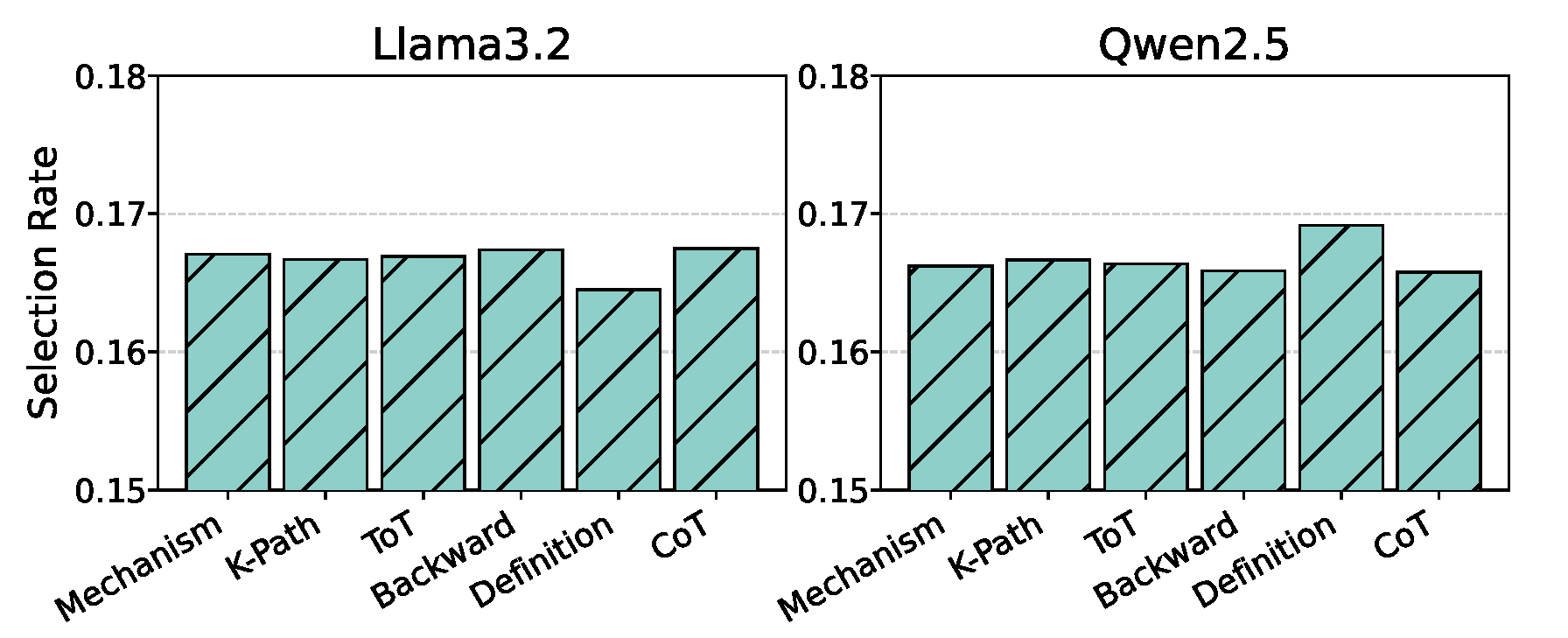}
    \caption{\small DrugBank.}
  \end{subfigure}\\[1mm]
  \begin{subfigure}[b]{0.46\textwidth}
    \centering
    \includegraphics[width=\linewidth]{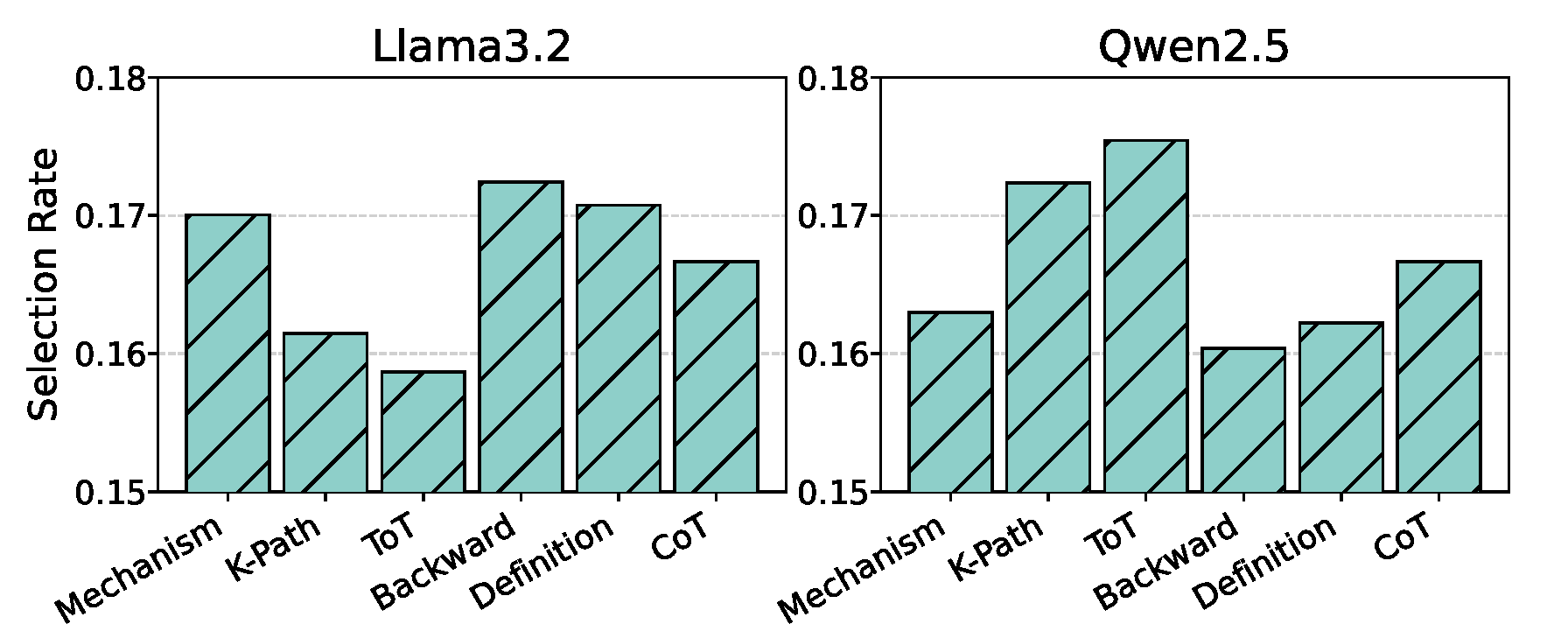}
    \caption{\small PharmaDB.}
  \end{subfigure}
  \caption{\small
    \textbf{Distribution of routed reasoning views across strategies.}
    The router maintains non-collapsed utilization of KG-derived and
    LLM-generated views, with dataset- and backbone-specific variation.
    }
  \label{fig:routing_comparison}
  \vspace{-3mm}
\end{figure}

\textbf{Router Analysis.}
Figure~\ref{fig:routing_comparison} analyzes how reasoning views are assigned across experts. The router does not collapse to a single reasoning strategy: KG-derived and LLM-generated views all receive non-trivial routing mass, with variations across datasets and backbones. This suggests that \textsc{DrugReason} keeps access to heterogeneous evidence sources rather than relying on one dominant prompt type. We view this analysis as a diagnostic of non-collapsed routing behavior; the ablation results in Table~\ref{tab:ablation_study} provide direct evidence that routing-based evidence assignment affects prediction.

\textbf{Hyper-parameter Sensitivity.}
Figure~\ref{fig:hyper} studies the effect of $\lambda_1$ and $\lambda_2$, which control routing entropy regularization and mutual distillation. \texttt{Llama3.2} remains relatively stable across settings, while \texttt{Qwen2.5} is more sensitive, especially to $\lambda_1$. This is consistent with the role of $\lambda_1$ in controlling expert assignment: overly weak or overly strong routing regularization can affect the balance between specialization and evidence sharing. The final training hyperparameters are listed in
Appendix~\ref{app:implementation}.


%% file: sections/app.tex
\appendix

\section{Prompt Templates and Input Filtering}
\label{app:prompt_filtering}

This appendix provides the prompt templates used to construct reasoning
views. These templates correspond to the reasoning strategies discussed in
Section~3.2. The gold label is never included in model inputs. For
teacher-generated rationales, we use the teacher LLM only during training
to construct mechanistic evidence views; the teacher is not queried during
inference. We apply deterministic filtering to remove empty outputs,
malformed responses, and template artifacts such as explicit label fields.
We do not rewrite or manually improve retained rationales.

\begin{table*}[t]
\centering
\scriptsize
\renewcommand{\arraystretch}{1.08}
\resizebox{\textwidth}{!}{
\begin{tabular}{p{0.17\textwidth}p{0.77\textwidth}}
\toprule
\textbf{View Type} & \textbf{Template} \\
\midrule
No reasoning &
\texttt{Given the candidate pair [ENTITY\_1] and [ENTITY\_2], predict the relation label for the target benchmark. Return only one valid label.} \\

Definition &
\texttt{Definition of [ENTITY\_1]: [DEF\_1]. Definition of [ENTITY\_2]: [DEF\_2]. Based on these biomedical definitions, predict the relation label for the candidate pair. Return only one valid label.} \\

K-Path &
\texttt{Candidate pair: [ENTITY\_1] and [ENTITY\_2]. Extracted biomedical paths: [PATH\_1]; [PATH\_2]; ... Based on these graph paths, predict the relation label. Return only one valid label.} \\

Mechanism &
\texttt{Focus on the biomedical mechanism connecting [ENTITY\_1] and [ENTITY\_2]. Describe relevant molecular targets, pathways, pharmacological processes, or therapeutic mechanisms. Do not restate the answer directly; justify the relationship mechanistically.} \\

Chain-of-Thought &
\texttt{Reason step by step about the possible biomedical relationship between [ENTITY\_1] and [ENTITY\_2]. Consider pharmacological effects, disease mechanisms, pathway connections, and interaction evidence.} \\

Tree-of-Thought &
\texttt{Generate multiple plausible biomedical hypotheses for the relationship between [ENTITY\_1] and [ENTITY\_2]. Compare these hypotheses and identify which one is most biologically plausible.} \\

Forward--Backward &
\texttt{First reason forward from [ENTITY\_1] to [ENTITY\_2]. Then reason backward from [ENTITY\_2] to [ENTITY\_1]. Check whether the two directions support a consistent biomedical relationship.} \\
\bottomrule
\end{tabular}
}
\caption{\small
\textbf{Prompt templates for reasoning-view construction.}
Placeholders are replaced with candidate entities, biomedical definitions,
or KG paths. Gold labels are omitted from all inputs.
}
\label{tab:app_prompt_templates}
\end{table*}

\paragraph{Filtering.}
After generation, we retain only non-empty, well-formed rationales. We
remove explicit label fields, answer-only artifacts, malformed outputs,
and responses that fail to follow the requested format. This filtering is
deterministic and does not involve rewriting the content of retained
rationales. This design ensures that LLM-generated views provide
mechanistic evidence rather than direct label leakage.

\section{Implementation Details}
\label{app:implementation}

We train \textsc{DrugReason} in two stages. In Stage 1, we optimize the
routed supervised prediction objective together with routing
regularization. In Stage 2, we keep the routed supervised objective and
add cross-expert mutual distillation. The coefficient $\lambda_1$ controls
routing load regularization, and $\lambda_2$ controls the mutual
distillation loss. We use $lr_{s1}$ and $lr_{s2}$ for expert training in
Stage 1 and Stage 2, respectively, and $lr_{misc}$ for non-expert
components such as the router.

\begin{table}[t]
\centering
\small
\renewcommand{\arraystretch}{1.05}
\begin{tabular}{lccc}
\toprule
\textbf{Hyperparameter} & \textbf{DDInter} & \textbf{DrugBank} & \textbf{PharmaDB} \\
\midrule
epochs & 1 & 1 & 1 \\
batch size & 8 & 8 & 8 \\
max length & 768 & 768 & 768 \\
$lr_{s1}$ & 3e-4 & 1e-4 & 5e-4 \\
$lr_{s2}$ & 5e-4 & 1e-4 & 1e-4 \\
$lr_{misc}$ & 1e-4 & 1e-4 & 1e-4 \\
$\lambda_2$ & 0.01 & 0.10 & 0.00 \\
$\lambda_1$ & 0.01 & 0.01 & 0.01 \\
$\tau$ & 1.0 & 1.0 & 1.0 \\
seed & 42 & 42 & 42 \\
\bottomrule
\end{tabular}
\caption{\small
\textbf{Llama3.2 training hyperparameters by dataset.}
$\lambda_1$ controls routing regularization and $\lambda_2$ controls
mutual distillation.
}
\label{tab:app_llama_hparams}
\end{table}

\begin{table}[t]
\centering
\small
\renewcommand{\arraystretch}{1.05}
\begin{tabular}{lccc}
\toprule
\textbf{Hyperparameter} & \textbf{DDInter} & \textbf{DrugBank} & \textbf{PharmaDB} \\
\midrule
epochs & 2 & 1 & 1 \\
batch size & 8 & 8 & 8 \\
max length & 768 & 768 & 768 \\
$lr_{s1}$ & 3e-5 & 1e-4 & 1e-4 \\
$lr_{s2}$ & 5e-5 & 1e-4 & 5e-4 \\
$lr_{misc}$ & 1e-4 & 1e-4 & 1e-4 \\
$\lambda_2$ & 0.10 & 0.01 & 0.10 \\
$\lambda_1$ & 0.01 & 0.01 & 0.01 \\
$\tau$ & 1.0 & 1.0 & 1.0 \\
seed & 42 & 42 & 42 \\
\bottomrule
\end{tabular}
\caption{\small
\textbf{Qwen2.5 training hyperparameters by dataset.}
$\lambda_1$ controls routing regularization and $\lambda_2$ controls
mutual distillation.
}
\label{tab:app_qwen_hparams}
\end{table}

\section{Additional Control Results}
\label{app:additional_results}

The main text reports compact control results to support RQ2. Here we
provide the full control tables. These controls are designed to test
whether the gains of \textsc{DrugReason} come merely from using more
evidence or from using KG structure alone. They use the same candidate
pairs and evaluation protocol as the main experiments.

\subsection{Static Prompt-Fusion Controls}
\label{app:prompt_controls}

Table~\ref{tab:app_prompt_controls} reports the full prompt-control
results. Single-prompt variants evaluate one reasoning view at a time.
Concat-all performs input-level fusion by placing all reasoning views into
one prompt. Majority vote and mean vote perform output-level fusion over
view-level predictions. These controls complement the compact results in
Table~\ref{tab:control_compact} of the main text.

\begin{table*}[t]
\centering
\scriptsize
\renewcommand{\arraystretch}{1.05}
\resizebox{\textwidth}{!}{
\begin{tabular}{l|cc|cc}
\toprule
\multirow{2}{*}{\textbf{Variant}} &
\multicolumn{2}{c|}{\textbf{DDInter}} &
\multicolumn{2}{c}{\textbf{PharmaDB}} \\
& \textbf{Acc.} & \textbf{Macro-F1} & \textbf{Acc.} & \textbf{Macro-F1} \\
\midrule
Concat-all             & 20.40 & 8.61  & 43.25 & 20.13 \\
Chain-of-Thought       & 3.80  & 6.03  & 41.27 & 22.33 \\
Definition             & 15.80 & 11.73 & 43.25 & 30.64 \\
Forward--Backward      & 20.50 & 14.88 & 42.86 & 29.24 \\
K-Path                 & 34.00 & 22.60 & 42.86 & 27.82 \\
K-Paths direct prompt   & 34.00 & 23.36 & 42.86 & 27.82 \\
Majority vote          & 18.40 & 11.55 & 42.46 & 23.24 \\
Mean vote              & 8.10  & 7.40  & 46.43 & 35.44 \\
Mechanism              & 4.60  & 5.15  & 46.03 & 34.52 \\
Tree-of-Thought        & 31.10 & 23.47 & 49.60 & 37.63 \\
\bottomrule
\end{tabular}
}
\caption{\small
\textbf{Detailed static prompt-fusion and single-prompt controls.}
Results are reported in percentage. These controls show that using more
reasoning views without adaptive routing does not reliably improve
performance.
}
\label{tab:app_prompt_controls}
\end{table*}

\subsection{Knowledge Graph Embedding Controls}
\label{app:kge_controls}

Table~\ref{tab:app_kge_controls} reports KGE baselines used to test
whether KG structure alone is sufficient without language-level
mechanistic reasoning. These results complement the compact KGE column in
Table~\ref{tab:control_compact} of the main text.

\begin{table}[t]
\centering
\small
\renewcommand{\arraystretch}{1.05}
\begin{tabular}{l|cc|cc}
\toprule
\multirow{2}{*}{\textbf{Dataset}} &
\multicolumn{2}{c|}{\textbf{TransE}} &
\multicolumn{2}{c}{\textbf{DistMult}} \\
& \textbf{Acc.} & \textbf{Macro-F1} & \textbf{Acc.} & \textbf{Macro-F1} \\
\midrule
DDInter  & 43.00 & 30.58 & 36.70 & 26.67 \\
DrugBank & 11.80 & 4.59  & 9.90  & 1.22  \\
PharmaDB & 37.70 & 36.43 & 32.94 & 32.05 \\
\bottomrule
\end{tabular}
\caption{\small
\textbf{Knowledge graph embedding baselines.}
Results are reported in percentage. These controls evaluate graph
structure alone, without language-level reasoning views or expert routing.
}
\label{tab:app_kge_controls}
\end{table}

\section{LLM Usage}

We used \texttt{GPT-4o} as a teacher model to generate training-time
mechanistic rationales, as described in the method. The teacher model is
not queried during inference. Separately, large language models were used
to assist with editing and polishing the manuscript, including improving
clarity, grammar, and readability. They were not used to generate
experimental results or alter reported numerical findings.

%% file: custom.bib
@article{k-path,
  title={K-paths: Reasoning over graph paths for drug repurposing and drug interaction prediction},
  author={Abdullahi, Tassallah and Gemou, Ioanna and Nayak, Nihal V and Murtaza, Ghulam and Bach, Stephen H and Eickhoff, Carsten and Singh, Ritambhara},
  journal={arXiv preprint arXiv:2502.13344},
  year={2025}
}

@article{Ji_2023,
   title={Survey of Hallucination in Natural Language Generation},
   volume={55},
   ISSN={1557-7341},
   url={http://dx.doi.org/10.1145/3571730},
   DOI={10.1145/3571730},
   number={12},
   journal={ACM Computing Surveys},
   publisher={Association for Computing Machinery (ACM)},
   author={Ji, Ziwei and Lee, Nayeon and Frieske, Rita and Yu, Tiezheng and Su, Dan and Xu, Yan and Ishii, Etsuko and Bang, Ye Jin and Madotto, Andrea and Fung, Pascale},
   year={2023},
   month=mar, pages={1–38} }

@article{perdomo2024knowledge,
  title={Knowledge Graphs for drug repurposing: a review of databases and methods},
  author={Perdomo-Quinteiro, Pablo and Belmonte-Hern{\'a}ndez, Alberto},
  journal={Briefings in Bioinformatics},
  volume={25},
  number={6},
  pages={bbae461},
  year={2024},
  publisher={Oxford University Press}
}

@article{chen2025benchmarking,
  title={Benchmarking large language models for biomedical natural language processing applications and recommendations},
  author={Chen, Qingyu and Hu, Yan and Peng, Xueqing and Xie, Qianqian and Jin, Qiao and Gilson, Aidan and Singer, Maxwell B and Ai, Xuguang and Lai, Po-Ting and Wang, Zhizheng and others},
  journal={Nature communications},
  volume={16},
  number={1},
  pages={3280},
  year={2025},
  publisher={Nature Publishing Group UK London}
}

@article{vaswani2017attention,
  title={Attention is all you need},
  author={Vaswani, Ashish and Shazeer, Noam and Parmar, Niki and Uszkoreit, Jakob and Jones, Llion and Gomez, Aidan N and Kaiser, {\L}ukasz and Polosukhin, Illia},
  journal={Advances in neural information processing systems},
  volume={30},
  year={2017}
}

@article{hamilton2017inductive,
  title={Inductive representation learning on large graphs},
  author={Hamilton, Will and Ying, Zhitao and Leskovec, Jure},
  journal={Advances in neural information processing systems},
  volume={30},
  year={2017}
}

@article{dwivedi2020generalization,
  title={A generalization of transformer networks to graphs},
  author={Dwivedi, Vijay Prakash and Bresson, Xavier},
  journal={arXiv preprint arXiv:2012.09699},
  year={2020}
}

@article{yang2014embedding,
  title={Embedding entities and relations for learning and inference in knowledge bases},
  author={Yang, Bishan and Yih, Wen-tau and He, Xiaodong and Gao, Jianfeng and Deng, Li},
  journal={arXiv preprint arXiv:1412.6575},
  year={2014}
}

@article{bordes2013translating,
  title={Translating embeddings for modeling multi-relational data},
  author={Bordes, Antoine and Usunier, Nicolas and Garcia-Duran, Alberto and Weston, Jason and Yakhnenko, Oksana},
  journal={Advances in neural information processing systems},
  volume={26},
  year={2013}
}

@article{han2022trusted,
  title={Trusted multi-view classification with dynamic evidential fusion},
  author={Han, Zongbo and Zhang, Changqing and Fu, Huazhu and Zhou, Joey Tianyi},
  journal={IEEE transactions on pattern analysis and machine intelligence},
  volume={45},
  number={2},
  pages={2551--2566},
  year={2022},
  publisher={IEEE}
}

@inproceedings{sahu2021adaptive,
  title={Adaptive fusion techniques for multimodal data},
  author={Sahu, Gaurav and Vechtomova, Olga},
  booktitle={Proceedings of the 16th conference of the European chapter of the Association for Computational Linguistics: Main Volume},
  pages={3156--3166},
  year={2021}
}

@article{liu2025generalist,
  title={A generalist medical language model for disease diagnosis assistance},
  author={Liu, Xiaohong and Liu, Hao and Yang, Guoxing and Jiang, Zeyu and Cui, Shuguang and Zhang, Zhaoze and Wang, Huan and Tao, Liyuan and Sun, Yongchang and Song, Zhu and others},
  journal={Nature medicine},
  volume={31},
  number={3},
  pages={932--942},
  year={2025},
  publisher={Nature Publishing Group US New York}
}

@article{singhal2023large,
  title={Large language models encode clinical knowledge},
  author={Singhal, Karan and Azizi, Shekoofeh and Tu, Tao and Mahdavi, S Sara and Wei, Jason and Chung, Hyung Won and Scales, Nathan and Tanwani, Ajay and Cole-Lewis, Heather and Pfohl, Stephen and others},
  journal={Nature},
  volume={620},
  number={7972},
  pages={172--180},
  year={2023},
  publisher={Nature Publishing Group UK London}
}

@inproceedings{abdullahi2025k,
  title={K-paths: Reasoning over graph paths for drug repurposing and drug interaction prediction},
  author={Abdullahi, Tassallah and Gemou, Ioanna and Nayak, Nihal V and Murtaza, Ghulam and Bach, Stephen H and Eickhoff, Carsten and Singh, Ritambhara},
  booktitle={Proceedings of the 31st ACM SIGKDD Conference on Knowledge Discovery and Data Mining V. 2},
  pages={5--16},
  year={2025}
}

@article{wei2024drugrealign,
  title={DrugReAlign: a multisource prompt framework for drug repurposing based on large language models},
  author={Wei, Jinhang and Zhuo, Linlin and Fu, Xiangzheng and Zeng, XiangXiang and Wang, Li and Zou, Quan and Cao, Dongsheng},
  journal={BMC biology},
  volume={22},
  number={1},
  pages={226},
  year={2024},
  publisher={Springer}
}

@article{pushpakom2019drug,
  title={Drug repurposing: progress, challenges and recommendations},
  author={Pushpakom, Sudeep and Iorio, Francesco and Eyers, Patrick A and Escott, K Jane and Hopper, Shirley and Wells, Andrew and Doig, Andrew and Guilliams, Tim and Latimer, Joanna and McNamee, Christine and others},
  journal={Nature reviews Drug discovery},
  volume={18},
  number={1},
  pages={41--58},
  year={2019},
  publisher={Nature Publishing Group}
}

@article{hopkins2008network,
  title={Network pharmacology: the next paradigm in drug discovery},
  author={Hopkins, Andrew L},
  journal={Nature chemical biology},
  volume={4},
  number={11},
  pages={682--690},
  year={2008},
  publisher={Nature Publishing Group US New York}
}

@article{drugrealign,
  title={DrugReAlign: a multisource prompt framework for drug repurposing based on large language models},
  author={Wei, Jinhang and Zhuo, Linlin and Fu, Xiangzheng and Zeng, XiangXiang and Wang, Li and Zou, Quan and Cao, Dongsheng},
  journal={BMC biology},
  volume={22},
  number={1},
  pages={226},
  year={2024},
  publisher={Springer}
}

@article{hgtdr,
  title={HGTDR: Advancing drug repurposing with heterogeneous graph transformers},
  author={Gharizadeh, Ali and Abbasi, Karim and Ghareyazi, Amin and Mofrad, Mohammad RK and Rabiee, Hamid R},
  journal={Bioinformatics},
  volume={40},
  number={7},
  pages={btae349},
  year={2024},
  publisher={Oxford University Press}
}

@article{DRMAHGC,
  title={Drug repositioning by multi-aspect heterogeneous graph contrastive learning and positive-fusion negative sampling strategy},
  author={Liu, Junkai and Hu, Fuyuan and Zou, Quan and Tiwari, Prayag and Wu, Hongjie and Ding, Yijie},
  journal={Information Fusion},
  volume={112},
  pages={102563},
  year={2024},
  publisher={Elsevier}
}

@article{medreason,
  title={Medreason: Eliciting factual medical reasoning steps in llms via knowledge graphs},
  author={Wu, Juncheng and Deng, Wenlong and Li, Xingxuan and Liu, Sheng and Mi, Taomian and Peng, Yifan and Xu, Ziyang and Liu, Yi and Cho, Hyunjin and Choi, Chang-In and others},
  journal={arXiv preprint arXiv:2504.00993},
  year={2025}
}

@article{TxGNN,
  title={A foundation model for clinician-centered drug repurposing},
  author={Huang, Kexin and Chandak, Payal and Wang, Qianwen and Havaldar, Shreyas and Vaid, Akhil and Leskovec, Jure and Nadkarni, Girish N and Glicksberg, Benjamin S and Gehlenborg, Nils and Zitnik, Marinka},
  journal={Nature Medicine},
  volume={30},
  number={12},
  pages={3601--3613},
  year={2024},
  publisher={Nature Publishing Group US New York}
}

@article{Rephetio,
  title={Systematic integration of biomedical knowledge prioritizes drugs for repurposing},
  author={Himmelstein, Daniel Scott and Lizee, Antoine and Hessler, Christine and Brueggeman, Leo and Chen, Sabrina L and Hadley, Dexter and Green, Ari and Khankhanian, Pouya and Baranzini, Sergio E},
  journal={elife},
  volume={6},
  pages={e26726},
  year={2017},
  publisher={eLife Sciences Publications, Ltd}
}

@article{sys_kg,
  title={Systematic integration of biomedical knowledge prioritizes drugs for repurposing},
  author={Himmelstein, Daniel Scott and Lizee, Antoine and Hessler, Christine and Brueggeman, Leo and Chen, Sabrina L and Hadley, Dexter and Green, Ari and Khankhanian, Pouya and Baranzini, Sergio E},
  journal={elife},
  volume={6},
  pages={e26726},
  year={2017},
  publisher={eLife Sciences Publications, Ltd}
}

@article{kg_review,
  title={Knowledge Graphs for drug repurposing: a review of databases and methods},
  author={Perdomo-Quinteiro, Pablo and Belmonte-Hern{\'a}ndez, Alberto},
  journal={Briefings in Bioinformatics},
  volume={25},
  number={6},
  pages={bbae461},
  year={2024},
  publisher={Oxford University Press}
}

@article{silico,
  title={Network-based in silico drug efficacy screening},
  author={Guney, Emre and Menche, J{\"o}rg and Vidal, Marc and Bar{\'a}basi, Albert-L{\'a}szl{\'o}},
  journal={Nature communications},
  volume={7},
  number={1},
  pages={10331},
  year={2016},
  publisher={Nature Publishing Group UK London}
}

@article{Med-PaLM,
  title={Large language models encode clinical knowledge},
  author={Singhal, Karan and Azizi, Shekoofeh and Tu, Tao and Mahdavi, S Sara and Wei, Jason and Chung, Hyung Won and Scales, Nathan and Tanwani, Ajay and Cole-Lewis, Heather and Pfohl, Stephen and others},
  journal={Nature},
  volume={620},
  number={7972},
  pages={172--180},
  year={2023},
  publisher={Nature Publishing Group}
}

@article{llm_med_benchmark,
  title={Benchmarking large language models for biomedical natural language processing applications and recommendations},
  author={Chen, Qingyu and Hu, Yan and Peng, Xueqing and Xie, Qianqian and Jin, Qiao and Gilson, Aidan and Singer, Maxwell B and Ai, Xuguang and Lai, Po-Ting and Wang, Zhizheng and others},
  journal={Nature communications},
  volume={16},
  number={1},
  pages={3280},
  year={2025},
  publisher={Nature Publishing Group UK London}
}

@article{graph2prompt,
  title={Biomedical knowledge graph-optimized prompt generation for large language models},
  author={Soman, Karthik and Rose, Peter W and Morris, John H and Akbas, Rabia E and Smith, Brett and Peetoom, Braian and Villouta-Reyes, Catalina and Cerono, Gabriel and Shi, Yongmei and Rizk-Jackson, Angela and others},
  journal={Bioinformatics},
  volume={40},
  number={9},
  pages={btae560},
  year={2024},
  publisher={Oxford University Press}
}

@article{BioGPT,
  title={BioGPT: generative pre-trained transformer for biomedical text generation and mining},
  author={Luo, Renqian and Sun, Liai and Xia, Yingce and Qin, Tao and Zhang, Sheng and Poon, Hoifung and Liu, Tie-Yan},
  journal={Briefings in bioinformatics},
  volume={23},
  number={6},
  pages={bbac409},
  year={2022},
  publisher={Oxford University Press}
}

@article{PubMedGPT,
  title={GPT meets PubMed: a novel approach to literature review using a large language model to crowdsource migraine medication},
  author={Mackenzie, Elyse and Cheng, Roger and Zhang, Pengfei},
  year={2025}
}

@article{MedFound,
  title={A generalist medical language model for disease diagnosis assistance},
  author={Liu, Xiaohong and Liu, Hao and Yang, Guoxing and Jiang, Zeyu and Cui, Shuguang and Zhang, Zhaoze and Wang, Huan and Tao, Liyuan and Sun, Yongchang and Song, Zhu and others},
  journal={Nature medicine},
  volume={31},
  number={3},
  pages={932--942},
  year={2025},
  publisher={Nature Publishing Group US New York}
}

@article{DrugAgent,
  title={Drugagent: Automating ai-aided drug discovery programming through llm multi-agent collaboration},
  author={Liu, Sizhe and Lu, Yizhou and Chen, Siyu and Hu, Xiyang and Zhao, Jieyu and Lu, Yingzhou and Zhao, Yue},
  journal={arXiv preprint arXiv:2411.15692},
  year={2024}
}

@article{Tx-LLM,
  title={Tx-llm: A large language model for therapeutics},
  author={Chaves, Juan Manuel Zambrano and Wang, Eric and Tu, Tao and Vaishnav, Eeshit Dhaval and Lee, Byron and Mahdavi, S Sara and Semturs, Christopher and Fleet, David and Natarajan, Vivek and Azizi, Shekoofeh},
  journal={arXiv preprint arXiv:2406.06316},
  year={2024}
}

@article{DrugMCTS,
  title={DrugMCTS: a drug repurposing framework combining multi-agent, RAG and Monte Carlo Tree Search},
  author={Yang, Zerui and Wan, Yuwei and Li, Yinqiao and Matsuda, Yudai and Xie, Tong and Song, Linqi},
  journal={arXiv preprint arXiv:2507.07426},
  year={2025}
}

@article{PharmAgents2025,
  title={Pharmagents: Building a virtual pharma with large language model agents},
  author={Gao, Bowen and Huang, Yanwen and Liu, Yiqiao and Xie, Wenxuan and Ma, Wei-Ying and Zhang, Ya-Qin and Lan, Yanyan},
  journal={arXiv preprint arXiv:2503.22164},
  year={2025}
}

@article{wei2022chain,
  title={Chain-of-thought prompting elicits reasoning in large language models},
  author={Wei, Jason and Wang, Xuezhi and Schuurmans, Dale and Bosma, Maarten and Xia, Fei and Chi, Ed and Le, Quoc V and Zhou, Denny and others},
  journal={Advances in neural information processing systems},
  volume={35},
  pages={24824--24837},
  year={2022}
}

@article{yao2023tree,
  title={Tree of thoughts: Deliberate problem solving with large language models},
  author={Yao, Shunyu and Yu, Dian and Zhao, Jeffrey and Shafran, Izhak and Griffiths, Tom and Cao, Yuan and Narasimhan, Karthik},
  journal={Advances in neural information processing systems},
  volume={36},
  pages={11809--11822},
  year={2023}
}

@article{xiong2022ddinter,
  title={DDInter: an online drug--drug interaction database towards improving clinical decision-making and patient safety},
  author={Xiong, Guoli and Yang, Zhijiang and Yi, Jiacai and Wang, Ningning and Wang, Lei and Zhu, Huimin and Wu, Chengkun and Lu, Aiping and Chen, Xiang and Liu, Shao and others},
  journal={Nucleic acids research},
  volume={50},
  number={D1},
  pages={D1200--D1207},
  year={2022},
  publisher={Oxford University Press}
}

@article{wishart2018drugbank,
  title={DrugBank 5.0: a major update to the DrugBank database for 2018},
  author={Wishart, David S and Feunang, Yannick D and Guo, An C and Lo, Elvis J and Marcu, Ana and Grant, Jason R and Sajed, Tanvir and Johnson, Daniel and Li, Carin and Sayeeda, Zinat and others},
  journal={Nucleic acids research},
  volume={46},
  number={D1},
  pages={D1074--D1082},
  year={2018},
  publisher={Oxford University Press}
}

@article{himmelstein2017systematic,
  title={Systematic integration of biomedical knowledge prioritizes drugs for repurposing},
  author={Himmelstein, Daniel Scott and Lizee, Antoine and Hessler, Christine and Brueggeman, Leo and Chen, Sabrina L and Hadley, Dexter and Green, Ari and Khankhanian, Pouya and Baranzini, Sergio E},
  journal={elife},
  volume={6},
  pages={e26726},
  year={2017},
  publisher={eLife Sciences Publications, Ltd}
}

@article{gcn,
  title={Semi-supervised classification with graph convolutional networks},
  author={Kipf, Thomas N and Welling, Max},
  journal={International Conference on Learning Representations},
  year={2017}
}

@article{gat,
  title={Graph attention networks},
  author={Veli{\v{c}}kovi{\'c}, Petar and Cucurull, Guillem and Casanova, Arantxa and Romero, Adriana and Lio, Pietro and Bengio, Yoshua},
  journal={International Conference on Learning Representations},
  year={2018}
}

@article{gin,
  title={How powerful are graph neural networks?},
  author={Xu, Keyulu and Hu, Weihua and Leskovec, Jure and Jegelka, Stefanie},
  journal={International Conference on Learning Representations},
  year={2019}
}

@article{gpt4o,
  title={GPT-4o System Card},
  author={OpenAI},
  journal={arXiv preprint arXiv:2410.21276},
  year={2024}
}

@article{llama3,
  title={The Llama 3 Herd of Models},
  author={Dubey, Abhimanyu and Jauhri, Abhinav and Pandey, Abhinav and others},
  journal={arXiv preprint arXiv:2407.21783},
  year={2024}
}

@article{qwen25,
  title={Qwen2.5 Technical Report},
  author={Yang, An and Hui, Binyuan and Zhang, Bubing and others},
  journal={arXiv preprint arXiv:2412.15115},
  year={2024}
}

@article{roberta,
  title={RoBERTa: A Robustly Optimized BERT Pretraining Approach},
  author={Liu, Yinhan and Ott, Myle and Goyal, Naman and Du, Jingfei and Joshi, Mandar and Chen, Danqi and Levy, Omer and Lewis, Mike and Zettlemoyer, Luke and Stoyanov, Veselin},
  journal={arXiv preprint arXiv:1907.11692},
  year={2019}
}

@article{chandak2023building,
  title={Building a knowledge graph to enable precision medicine},
  author={Chandak, Payal and Huang, Kexin and Zitnik, Marinka},
  journal={Scientific Data},
  volume={10},
  number={1},
  pages={67},
  url={https://doi.org/10.1038/s41597-023-01960-3},
  year={2023},
  publisher={Nature Publishing Group}
}

@misc{soman2024biomedicalknowledgegraphoptimizedprompt,
      title={Biomedical knowledge graph-optimized prompt generation for large language models}, 
      author={Karthik Soman and Peter W Rose and John H Morris and Rabia E Akbas and Brett Smith and Braian Peetoom and Catalina Villouta-Reyes and Gabriel Cerono and Yongmei Shi and Angela Rizk-Jackson and Sharat Israni and Charlotte A Nelson and Sui Huang and Sergio E Baranzini},
      year={2024},
      eprint={2311.17330},
      archivePrefix={arXiv},
      primaryClass={cs.CL},
      url={https://arxiv.org/abs/2311.17330}, 
}

@misc{cabello2025megmedicalknowledgeaugmentedlarge,
      title={MEG: Medical Knowledge-Augmented Large Language Models for Question Answering}, 
      author={Laura Cabello and Carmen Martin-Turrero and Uchenna Akujuobi and Anders Søgaard and Carlos Bobed},
      year={2025},
      eprint={2411.03883},
      archivePrefix={arXiv},
      primaryClass={cs.CL},
      url={https://arxiv.org/abs/2411.03883}, 
}

@misc{li2024biomedragretrievalaugmentedlarge,
      title={BiomedRAG: A Retrieval Augmented Large Language Model for Biomedicine}, 
      author={Mingchen Li and Halil Kilicoglu and Hua Xu and Rui Zhang},
      year={2024},
      eprint={2405.00465},
      archivePrefix={arXiv},
      primaryClass={cs.CL},
      url={https://arxiv.org/abs/2405.00465}, 
}

@misc{sohn2025rationaleguidedretrievalaugmentedgeneration,
      title={Rationale-Guided Retrieval Augmented Generation for Medical Question Answering}, 
      author={Jiwoong Sohn and Yein Park and Chanwoong Yoon and Sihyeon Park and Hyeon Hwang and Mujeen Sung and Hyunjae Kim and Jaewoo Kang},
      year={2025},
      eprint={2411.00300},
      archivePrefix={arXiv},
      primaryClass={cs.CL},
      url={https://arxiv.org/abs/2411.00300}, 
}

@inproceedings{duan2024shifting,
  title={Shifting attention to relevance: Towards the predictive uncertainty quantification of free-form large language models},
  author={Duan, Jinhao and Cheng, Hao and Wang, Shiqi and Zavalny, Alex and Wang, Chenan and Xu, Renjing and Kailkhura, Bhavya and Xu, Kaidi},
  booktitle={Proceedings of the 62nd Annual Meeting of the Association for Computational Linguistics (Volume 1: Long Papers)},
  pages={5050--5063},
  year={2024}
}

@article{duan2024gtbench,
  title={Gtbench: Uncovering the strategic reasoning capabilities of llms via game-theoretic evaluations},
  author={Duan, Jinhao and Zhang, Renming and Diffenderfer, James and Kailkhura, Bhavya and Sun, Lichao and Stengel-Eskin, Elias and Bansal, Mohit and Chen, Tianlong and Xu, Kaidi},
  journal={Advances in Neural Information Processing Systems},
  volume={37},
  pages={28219--28253},
  year={2024}
}

@inproceedings{duan2023diffusion,
  title={Are diffusion models vulnerable to membership inference attacks?},
  author={Duan, Jinhao and Kong, Fei and Wang, Shiqi and Shi, Xiaoshuang and Xu, Kaidi},
  booktitle={International Conference on Machine Learning},
  pages={8717--8730},
  year={2023},
  organization={PMLR}
}
